\documentclass[11pt,a4paper]{article}
\usepackage[hyperref]{emnlp2018}
\usepackage{times}
\usepackage{diagbox}
\usepackage{latexsym}
\usepackage{graphicx}
\usepackage{multirow}
\usepackage{makecell}

\usepackage{url}

\aclfinalcopy 

\newcommand{\personachat}{\textsc{Persona-chat}}
\newcommand{\reddit}{\textsc{Reddit}}
\newcommand{\attentionmodel}{Transformer}

\title{Training Millions of Personalized Dialogue Agents}

\author{Pierre-Emmanuel Mazar\'e, Samuel Humeau, Martin Raison, Antoine Bordes \\
  Facebook \\
  {\tt \{pem, samuelhumeau, raison, abordes\}@fb.com }}

\date{}

\begin{document}
\maketitle
\begin{abstract}
Current dialogue systems are not very engaging for users, especially when trained end-to-end without relying on proactive reengaging scripted strategies.
\citet{personachat} showed that the engagement level of end-to-end dialogue models increases when conditioning them on text personas providing some personalized back-story to the model. 
However, the dataset used in \citep{personachat} is synthetic and of limited size as it contains around 1k different personas.
In this paper we introduce a new dataset providing 5 million personas and 700 million persona-based dialogues. Our experiments show that, at this scale, training using personas still improves the performance of end-to-end systems. In addition, we show that other tasks benefit from the wide coverage of our dataset by fine-tuning our model on the data from \citep{personachat} and achieving state-of-the-art results.
\end{abstract}

\section{Introduction}
End-to-end dialogue systems, based on neural architectures like bidirectional LSTMs or Memory Networks \cite{sukhbaatar2015end} trained directly by gradient descent on dialogue logs, have been showing promising performance in multiple contexts \cite{wen2016network,serban2016building,bordes2016learning}. One of their main advantages is that they can rely on large data sources of existing dialogues to learn to cover various domains without requiring any expert knowledge.
However, the flip side is that they also exhibit limited engagement, especially in chit-chat settings: they lack consistency and do not leverage proactive engagement strategies as (even partially) scripted chatbots do.

\citet{personachat} introduced the \personachat{} dataset as a solution to cope with this issue. This dataset consists of dialogues between pairs of agents with text profiles, or {\it personas}, attached to each of them. As shown in their paper, conditioning an end-to-end system on a given persona improves the engagement of a dialogue agent. This paves the way to potentially end-to-end personalized chatbots because the personas of the bots, by being short texts, could be easily edited by most users.
However, the \personachat{} dataset was created using an artificial data collection mechanism based on Mechanical Turk. As a result, neither dialogs nor personas can be fully representative of real user-bot interactions and the dataset coverage remains limited, containing a bit more than 1k different personas.

In this paper, we build a very large-scale persona-based dialogue dataset using conversations previously extracted from \reddit\footnote{\url{https://www.reddit.com/r/datasets/comments/3bxlg7/}}. With simple heuristics, we create a corpus of over 5 million personas spanning more than 700 million conversations.
We train persona-based end-to-end dialogue models on this dataset. These models outperform their counterparts that do not have access to personas, confirming results of \citet{personachat}. In addition, the coverage of our dataset seems very good since pre-training on it also leads to state-of-the-art results on the \personachat{} dataset.

\section{Related work}

With the rise of end-to-end dialogue systems, personalized trained systems have started to appear.
\citet{li2016} proposed to learn latent variables representing each speaker's bias/personality in a dialogue model. Other classic strategies include extracting explicit variables from structured knowledge bases or other symbolic sources as in \cite{ghazvininejad2017,joshi2017personalization,young2017augmenting}. Still, in the context of personal chatbots, it might be more desirable to condition on data that can be generated and interpreted by the user itself such as text rather than relying on some knowledge base facts that might not exist for everyone or a great variety of situations.
\personachat{} \cite{personachat} recently introduced a dataset of conversations revolving around human habits and preferences. In their experiments, they showed that conditioning on a text description of each speaker's habits, their \emph{persona}, improved dialogue modeling.

In this paper, we use a pre-existing \reddit{} data dump as data source. \reddit{} is a massive online message board. \citet{dodge2015evaluating} used it to assess chit-chat qualities of generic dialogue models. \citet{yang2018} used response prediction on \reddit{} as an auxiliary task in order to improve prediction performance on natural language inference problems.

\section{Building a dataset of millions of persona-based dialogues}

Our goal is to learn to predict responses based on a persona for a large variety of personas. To that end, we build a dataset of examples of the following form using data from \reddit:
\begin{itemize}
\itemsep0em
\item
Persona: [\textit{``I like sport"}, \textit{``I work a lot"}]
\item
Context: \textit{``I love running."}
\item
Response: \textit{``Me too! But only on weekends."}
\end{itemize}

The persona is a set of sentences representing the personality of the responding agent, the context is the utterance that it responds to, and the response is the answer to be predicted.

\subsection{Preprocessing}

As in \cite{dodge2015evaluating}, we use a preexisting dump of \reddit{}{} that consists of 1.7 billion comments. We tokenize sentences by padding all special characters with a space and splitting on whitespace characters. We create a dictionary containing the 250k most frequent tokens. We truncate comments that are longer than 100 tokens.

\subsection{Persona extraction}
\label{personaextraction}
We construct the persona of a user by gathering all the comments they wrote, splitting them into sentences, and selecting the sentences that satisfy the following rules: (i) each sentence must contain between 4 and 20 words or punctuation marks, (ii) it contains either the word \emph{I} or \emph{my}, (iii) at least one verb, and (iv) at least one noun, pronoun or adjective.

To handle the quantity of data involved, we limit the size of a persona to $N$ sentences for each user. We compare four different setups for persona creation. In the \emph{rules} setup, we select up to $N$ random sentences that satisfy the rules above. In the \emph{rules + classifier} setup, we filter with the rules then score the resulting sentences using a bag-of-words classifier that is trained to discriminate \personachat{} persona sentences from random comments. We manually tune a threshold on the score in order to select sentences. If there are more than $N$ eligible persona sentences for a given user, we keep the highest-scored ones. In the \emph{random from user} setup, we randomly select sentences uttered by the user while keeping the sentence length requirement above (we ignore the other rules). The \emph{random from dataset} baseline refers to random sentences from the dataset. They do not necessarily come from the same user. This last setup serves as a control mechanism to verify that the gains in prediction accuracy are due to the user-specific information contained in personas.

In the example at the beginning of this section, the response is clearly consistent with the persona. There may not always be such an obvious relationship between the two: the discussion topic may not be covered by the persona, a single user may write contradictory statements, and due to errors in the extraction process, some persona sentences may not represent a general trait of the user (e.g. \textit{I am feeling happy today}).

\subsection{Dataset creation\label{section:dataset}}
We take each pair of successive comments in a thread to form the context and response of an example. The persona corresponding to the response is extracted using one of the methods of Section~\ref{personaextraction}.
We split the dataset randomly between training, validation and test. Validation and test sets contain 50k examples each. We extract personas using training data only: test set responses cannot be contained explicitly in the persona.

In total, we select personas covering 4.6m users in the rule-based setups and 7.2m users in the random setups. This is a sizable fraction of the total 13.2m users of the dataset; depending on the persona selection setup, between 97 and 99.4\,\% of the training set examples are linked to a persona.

\section{End-to-end dialogue models}

\begin{figure*}
\center
\includegraphics[width=0.7\textwidth]{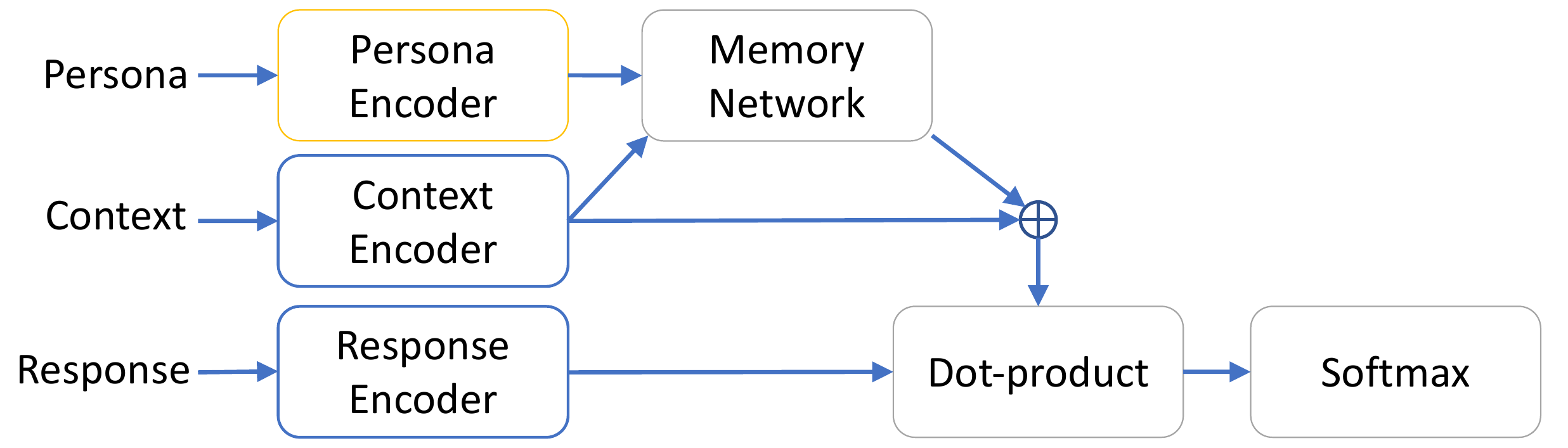}
\caption{\label{fig:personabased}Persona-based network architecture.}
\end{figure*}

We model dialogue by next utterance retrieval \cite{lowe2016evaluation}, where a response is picked among a set of candidates and not generated.

\subsection{Architecture}

The overall architecture is depicted in Fig. \ref{fig:personabased}. We encode the persona and the context using separate modules. As in \citet{personachat}, we combine the encoded context and persona using a 1-hop memory network with a residual connection, using the context as query and the set of persona sentences as memory. 
We also encode all candidate responses and compute the dot-product between all those candidate representations and the joint representation of the context and the persona. 
The predicted response is the candidate that maximizes the dot product.

We train by passing all the dot products through a softmax and maximizing the log-likelihood of the correct responses.
We use mini-batches of training examples and, for each example therein, all the responses of the other examples of the same batch are used as negative responses.

\subsection{Context and response encoders}

Both context and response encoders share the same architecture and word embeddings but have different weights in the subsequent layers.  We train three different encoder architectures.

\paragraph{Bag-of-words} applies two linear projections separated by a $\mathrm{tanh}$ non-linearity to the word embeddings. We then sum the resulting sentence representation across all positions in the sentence and divide the result by $\sqrt{n}$ where $n$ is the length of the sequence.

\paragraph{LSTM} applies a 2-layer bidirectional LSTM. We use the last hidden state as encoded sentence.

\paragraph{\attentionmodel{}}  is a variation of an End-to-end Memory Network \cite{sukhbaatar2015end} introduced by \citet{vaswani2017}. Based solely on attention mechanisms, it exhibited state-of-the-art performance on next utterance retrieval tasks in dialogues \cite{yang2018}. Here we use only its encoding module. We subsequently average the resulting representation across all positions in the sentence, yielding a fixed-size representation. 

\subsection{Persona encoder}
The persona encoder encodes each persona sentence separately. It relies on the same word embeddings as the context encoder and applies a linear layer on top of them. We then sum the representations across the sentence.

We deliberately choose a simpler architecture than the other encoders for performance reasons as the number of personas encoded for each batch is an order of magnitude greater than the number of training examples. Most personas are short sentences; we therefore expect a bag-of-words representation to encode them well.

\section{Experiments}

\begin{table}[t]
\begin{center}
\begin{tabular}{l|c|ccc}
& & \multicolumn{3}{c}{hits@k} \\
& Persona & k=1 & k=3 & k=10 \\
\hline
IR Baseline & No & 5.6 & 9.9 & 19.5 \\
\hline
BOW & No & 51.7 & 64.7 & 77.9 \\
BOW & Yes & 53.9 & 67.9 & 81.9 \\
LSTM & No & 63.1 & 75.6 & 87.3 \\
LSTM & Yes & 66.3	& 79.5 &	90.6 \\
\attentionmodel{} & No & 69.1 & 80.7 & 90.7 \\
\attentionmodel{} & Yes & \bf 74.4 & \bf 85.6 & \bf 94.2 \\

\end{tabular}
\caption{\label{tab:results} Test results when classifying the correct answer among a total of 100 possible answers.}
\end{center}
\end{table}

\begin{table*}[t]
\begin{center}
\begin{tabular*}{\textwidth}{l|l}
\it Context \it (Persona) & \it Predicted Answer \\
\hline
\it Where do you come from? & \\
\it \makecell[tl]{(I was born in London.)} & \it  \makecell[tl]{I'm from London, studying in Scotland.} \\
\it \makecell[tl]{(I was born in New York.)} & \it  \makecell[tl]{I'm from New York.} \\
\hline
\it What do you do? & \\
\it \makecell[tl]{(I am a doctor.) }& \it \makecell[tl]{I am a sleep and respiratory therapist.} \\
\it \makecell[tl]{(I am an engineer.)} & \it \makecell[tl]{I am a software developer.} \\

\end{tabular*}
\caption{\label{tab:examples} Sample predictions from the best model. In all selected cases the persona consists of a single sentence. The answer is constrained to be at most 10 tokens and is retrieved among 1M candidates sampled randomly from the training set.}
\end{center}
\end{table*}

We train models on the persona-based dialogue dataset described in Section \ref{section:dataset} and we evaluate its accuracy both on the original task and when transferring onto \personachat{}. 

\subsection{Experimental details}
We optimize network parameters using Adamax with a learning rate of $8\mathrm{e}{-4}$ on mini-batches of size 512. We initialize embeddings with FastText word vectors and optimize them during learning.

\paragraph{\reddit{}} LSTMs use a hidden size of 150; we concatenate the last hidden states for both directions and layers, resulting in a final representation of size 600. \attentionmodel{} architectures on reddit use 4 layers with a hidden size of 300 and 6 attention heads, resulting in a final representation of size 300. We use Spacy for part-of-speech tagging in order to verify the persona extraction rules. We distribute the training by splitting each batch across 8 GPUs; we stop training after 1 full epoch, which takes about 3 days.

\paragraph{\personachat{}} We used the revised version of the dataset where the personas have been rephrased, making it a harder task. The dataset being only a few thousands samples, we had to reduce the architecture to avoid overfitting for the models trained purely on \personachat{}. 2 layers, 2 attention heads, a dropout of 0.2 and keeping the size of the word embeddings to 300 units yield the highest accuracy on the validation set.

\paragraph{IR Baseline} As basic baseline, we use an information retrieval (IR) system that ranks candidate responses according to a TF-IDF weighted exact-match similarity with the context alone.

\subsection{Results}

\paragraph{Impact of personas}
We report the accuracy of the different architectures on the reddit task in Table~\ref{tab:results}. Conditioning on personas improves the prediction performance regardless of the encoder architecture. 
Table \ref{tab:examples} gives some examples of how the persona affects the predicted answer.

\begin{table}[t]
\begin{center}
\begin{tabular}{c|c|c}
$N$ & Persona selection & hits@1 \\
\hline
0 & -- & 69.1  \\
20 & rules + classifier & 70.7  \\
20 & rules & 71.3  \\
100 & rules + classifier & 72.5 \\
100 & rules & \bf 74.4 \\
100 & random from user & 73.8  \\
100 & random from dataset & 66.9  \\
\end{tabular}
\caption{\label{tab:personachoice} Retrieval precision on the \reddit{} test set using a \attentionmodel{} and different persona selection systems. $N$: maximum number of sentences per persona.}
\end{center}
\end{table}

\paragraph{Influence of the persona extraction}
In Table~\ref{tab:personachoice}, we report precision results for several persona extraction setups. The \textit{rules} setup improves the results somewhat, however adding the persona classifier actually degrades the results.
A possible interpretation is that the persona classifier is trained only on the \personachat{} revised personas, and that this selection might be too narrow and lack diversity.
Increasing the maximum persona size also improves the prediction performance.

\paragraph{Transfer learning}

\begin{table}[t]
\begin{center}
\begin{tabular}{l|cc}
&  \multicolumn{2}{c}{Validation set} \\
Training set & \personachat{} & \reddit{}\\
\hline
\personachat{} & 42.1 & 3.04 \\
\reddit{} &  25.6 & \bf 74.4  \\
FT-PC & \bf 60.7 & 65.5  \\
\hline
IR Baseline & 20.7 & 5.6 \\
\small{\cite{personachat}} & 35.4 & -- \\
\end{tabular}
\caption{\label{tab:transfer} hits@1 results for the best found \attentionmodel{} architecture on different test sets. FT-PC: \reddit{}-trained model fine-tuned on the \personachat{} training set. To be comparable to the state of the art on each dataset, results on \personachat{} are computed using 20 candidates, while results on \reddit{} use 100.} 
\end{center}
\end{table}

We compare the performance of transformer models trained on \reddit{} and on \personachat{} on both datasets. We report results in Table \ref{tab:transfer}. This architecture provides a strong improvement over the results of \cite{personachat}, jumping from 35.4\% hits@1 to 42.1\%. Pretraining the model on \reddit{} and then fine-tuning on \personachat{} pushes this score to 60.7\%, largely improving the state of the art. As expected, fine-tuning on \personachat{} reduces the performance on \reddit{}. However, directly testing on \personachat{} the model trained on \reddit{} without fine-tuning yields a very low result. This could be a consequence of a discrepancy between the style of personas of the two datasets.

\section{Conclusion}

This paper shows how to create a very large dataset for persona-based dialogue. We show that training models to align answers both with the persona of their author and the context improves the predicting performance. The trained models show promising coverage as exhibited by the state-of-the-art transfer results on the \personachat{} dataset. As pretraining leads to a considerable improvement in performance, future work could be done fine-tuning this model for various dialog systems. Future work may also entail building more advanced strategies to select a limited number of personas for each user while maximizing the prediction performance.

\bibliographystyle{acl_natbib_nourl}
\bibliography{persona_reddit}

\end{document}